\documentclass{article}

\usepackage{PRIMEarxiv}

\usepackage[utf8]{inputenc} 
\usepackage[T1]{fontenc}    
\usepackage{hyperref}       
\usepackage{url}            
\usepackage{booktabs}       
\usepackage{amsfonts}       
\usepackage{nicefrac}       
\usepackage{microtype}      
\usepackage{lipsum}
\usepackage{fancyhdr}       
\usepackage{graphicx}       
\graphicspath{{media/}}     

\title{Between Predictability and Randomness: Seeking Artistic Inspiration from AI Generative Models\thanks{This paper was presented as a keynote at the 50\textsuperscript{th} Linguistic Association of Canada and the United States (LACUS) conference in July 2024 and will be published in LACUS Forum 50.} 
}

\author{
  Olga Vechtomova \\
  University of Waterloo \\
  Waterloo, ON, Canada\\
  \texttt{ovechtom@uwaterloo.ca} \\
}

\begin{document}
\maketitle

\begin{abstract}
Artistic inspiration often emerges from language that is open to interpretation. This paper explores the use of AI-generated poetic lines as stimuli for creativity. Through analysis of two generative AI approaches—lines generated by Long Short-Term Memory Variational Autoencoders (LSTM-VAE) and complete poems by Large Language Models (LLMs)—I demonstrate that LSTM-VAE lines achieve their evocative impact through a combination of resonant imagery and productive indeterminacy. While LLMs produce technically accomplished poetry with conventional patterns, LSTM-VAE lines can engage the artist through semantic openness, unconventional combinations, and fragments that resist closure. Through the composition of an original poem, where narrative emerged organically through engagement with LSTM-VAE generated lines rather than following a predetermined structure, I demonstrate how these characteristics can serve as evocative starting points for authentic artistic expression.
\end{abstract}


\section{Introduction}
Artistic inspiration is a complex phenomenon that often emerges from the interplay between predictable and unpredictable elements. In the context of the Seed phase of creativity, where initial ideas spark the creative process~\cite{rubin2023creative}, artists have long employed techniques to disrupt conventional patterns and create new possibilities for expression. With the recent advances in artificial intelligence (AI) research, new opportunities arise for generating such creative stimuli through neural text generative models.
This paper investigates how AI-generated poetic lines can serve as sources of inspiration during the Seed phase, focusing particularly on their potential to balance structure and unpredictability. I examine two different approaches to the use of AI in poetry creation: individual lines generated by Long Short-Term Memory Variational Autoencoders (LSTM-VAE) that serve as creative starting points for human authors, leading to emergent narratives and original works, and complete poems autonomously generated by Large Language Models (LLMs). Through analysis of two poems—one that emerged through engagement with LSTM-VAE generated lines, and another by an LLM—I demonstrate how LSTM-VAE are particularly effective as sources of creative inspiration.
LSTM-VAE models, due to their design and training objectives, generate text that occupies a productive middle ground between randomness and predictability. Their output combines semantic openness with unconventional linguistic patterns, enabling artists to discover and develop original poetic narratives through genuine emotional engagement with the generated lines. In contrast, while LLMs produce technically accomplished poetry, their adherence to conventional forms tends to constrain rather than encourage creative interpretation. The contributions of this paper are threefold:

\begin{enumerate}

\item An examination of how AI can extend historical artistic techniques of creative disruption, establishing a framework for understanding machine-generated text as a source of artistic inspiration.

\item A comparative analysis of LSTM-VAE and LLM approaches to poetry generation, demonstrating how their different architectures influence their potential for inspiring creativity.

\item Evidence for how specific qualities of AI-generated lines—particularly their balance of structure and unpredictability—can enable the emergence of authentic artistic narratives~\footnote{In this paper, I use ``narrative'' in a broad sense: not merely as a traditional storyline with distinct characters and events, but as an emergent trajectory of meaning or emotional arc that arises from the arrangement of poetic lines.}.
\end{enumerate}

This investigation reveals how specific qualities of AI-generated lines—their linguistic patterns, semantic openness, and capacity for novel combinations—influence their potential as creative stimuli. The organic emergence of poetic narrative through engagement with LSTM-VAE lines demonstrates that productive linguistic innovation, rather than mere technical accomplishment, can inspire artistic process. These findings have important implications for developing AI systems that can effectively support human creativity.

\section{Inspiration and the Creative Process}
\label{sec:Inspiration}

The creative process in artistic endeavors, particularly in poetry and lyric writing, involves distinct phases that require different cognitive and artistic approaches. Rubin~\cite{rubin2023creative} formulates three key stages:

\textbf{Seed Phase.} This initial stage represents a state of creative receptivity where diverse stimuli can spark artistic inspiration. It involves heightened sensitivity to linguistic input (words, phrases, syntactic structures), openness to unexpected associations, and engagement with ambiguous elements that can trigger multiple interpretations.

\textbf{Experimentation.} In this intermediate phase, artists combine and reconfigure elements discovered during the Seed phase, engaging both generative and evaluative aspects of creative thinking. This phase represents a convergence of the initial inspiration and artistic vision, where raw elements take coherent form.

\textbf{Crafting.} The final stage involves applying technical skill and artistic judgment to refine the work. This requires both mastery of technique and sensitivity to the original creative impulse that emerged during the Seed phase.

While these stages are conceptually distinct, they can overlap in practice, with creative inspiration potentially emerging throughout the process. This study focuses specifically on the Seed phase and how AI-generated lines can serve as effective creative stimuli during this crucial initial stage. I argue that certain qualities of generated text—particularly unpredictability and semantic openness—can be especially valuable during the Seed phase, where disruption of habitual patterns often leads to creative breakthrough.

The effectiveness of unpredictable or indeterminate elements during the Seed phase aligns with established artistic practices throughout history. As will be discussed in the next section, many 20th-century artists, poets, and writers deliberately employed techniques that introduced randomness and unpredictability into their creative process, particularly during the initial stages of artistic creation.

This historical context helps us understand why certain characteristics of AI-generated lines— such as syntactic estrangement, deliberate incompleteness, and unconventional semantic relationships—might make them particularly effective as seeds for artistic inspiration. Just as traditional techniques like the cut-up method created space for new artistic possibilities, AI-generated lines with the right balance of unpredictability and coherence can serve as valuable starting points for creative exploration.

\section{Unpredictability in the Creative Process}
\label{sec:Unpredictability}

Artists have long sought methods to break free from conventional language patterns that can restrict creative expression. The cut-up technique, originally proposed by the Dadaists in the early 20th century, was one such method. It consisted of literally using scissors to cut up a page of printed text into words and letting them fall in random order. While the Dadaists intended this as a statement challenging the very notion of art, William Burroughs later popularized it as a literary technique. Burroughs argued that language is a ``lock'' that restricts creativity and confines us to predictable patterns~\cite{burroughs2012}. Musicians like David Bowie~\cite{jones2017david} and Kurt Cobain~\cite{cross2001heavier} adopted this technique in their songwriting. Bowie described using cut-ups to stimulate his imagination through random juxtaposition of concepts, stating: ``What I've used it for, more than anything else, is igniting anything that might be in my imagination.''~\cite{BowieCrackedActor}.

Syntactic defamiliarization—the deliberate disruption of expected syntactic patterns—emerged as another powerful technique for achieving creative unpredictability. Russian Formalist Viktor Shklovsky's concept of "ostranenie" ("estrangement" or "defamiliarization") provides a theoretical framework for understanding how breaking conventional syntactic-semantic relationships can make language feel fresh and strange~\cite{shklovsky1965art}. When artists deliberately create syntactic ``errors'' or incomplete structures, they force readers to engage with language in new ways, much like the cut-up technique. This syntactic estrangement prevents automatic processing of language and demands active meaning-making from the audience.

The intentional use of randomness and unpredictability extended beyond literature into visual art of the early to mid 20th century, with artists like Jean Arp, Georg Nees, Kurt Schwitters, and Vera Molnar incorporating random elements into their work. Georg Nees, considered to be the author of the first computer art, noted that randomness alone cannot be the source of art—there must be an aesthetic structure to narrow down the outcomes and produce discernible aesthetic information~\cite{nees1969generative}. Molnar observed that computer programs with random generators help artists explore a broader range of possibilities in finding the right artistic form~\cite{MolnarVimeo2019}.

Kurt Schwitters experimented with word unpredictability in his Merz-poems through collage techniques, incorporating various elements from different sources. His deliberate linguistic disruptions aimed to heighten emotional impact, as he explained regarding his poem "An Anna Blume": ```I love thou' is more expressive than `I love thee', because it's unusual. Had I written `I love thee' nobody would have noticed that I love Anna Blume''~\cite{schwitters2021myself}.

This tradition of intentional language disruption continued with the ``Language poetry'' movement of the 1970s, which posited that language dictates meaning rather than the opposite. Poets like Bob Perelman deliberately broke up language to engage readers in actively constructing meaning. More recently, novelist Tom Spanbauer's ``burnt tongue'' technique—deliberately misusing language to make it fresh—echoes these earlier approaches. As his student Chuck Palahniuk explains, it is ``a way of saying something, but saying it wrong, twisting it to slow down the reader''~\cite{palahniuk2020consider}. Eye tracking studies validate this approach, showing that predictable words are indeed skipped more often by readers and elicit shorter eye fixations. For example, readers spend less time on ``cake'' in ``the baker rushed the wedding cake'' than on ``pies'' in ``the baker rushed the wedding pies.''~\cite{Lowder2018}

\section{Predictability in Art}
Predictability in art has served both as a target of critique and as a deliberate artistic device. Artists have often employed predictable elements, triviality, and cliché as part of intentional artistic statements, using familiarity itself as a tool for engaging audience expectations.

The pop-art movement, which emerged in the 1950s and 1960s, is one of the best-known examples of an art movement using predictability and banality as a device, often with an intent to critique the superficiality of consumer culture. Artists such as Andy Warhol and Barbara Kruger often used recognizable imagery from advertisements, comics and popular culture with intentionally cliché and banal statements, such as ``I shop therefore I am''~\cite{Kruger1987}.

The conceptual art movement of the 1960s further explored predictability as an artistic device. In conceptual art, where the concept behind the work is prioritized over its aesthetic or material qualities, banality became a tool for challenging viewers' perceptions of art, language, and everyday life. Jenny Holzer's ``Truisms''~\cite{Holzer1978} exemplifies this approach, using everyday language and declarative sentences to create intentionally trivial-sounding statements like ``A little knowledge goes a long way'' or ``Action causes more trouble than thought.''

While Pop-art and Conceptual art deliberately employed predictability and banality as artistic devices, in literature, predictability is often identified as inauthentic and a form of artistic shortcut. This perspective is captured in Tom Spanbauer's concept of `received text.' As  Palahniuk explains: ``To Tom (Spanbauer), `wringing your hands' would be called `Received Text.' Like a cliché, but more subtle. The phrase might not be as bad as `warm as a summer's day' or `pretty as a picture', but the phrase was still a short-cut, or pathway well-worn by a lot of writers and easy to feed into a story"~\cite{palahniuk2020consider}.

\section{Complexity and Aesthetic Perception}
The relationship between aesthetic pleasure and complexity, known as the inverted U-shaped or ``Wundt'' effect, suggests that stimuli of intermediate complexity are most pleasing. This effect, first explored by Berlyne~\cite{Berlyne1975}, has been observed in various domains such as music~\cite{Gold9397} and product design~\cite{Althuizen2021}.

Complexity is typically expressed in features such as predictability, surprise and uncertainty. Gold et al~\cite{Gold9397} offered a possible explanation for the inverted U-shaped effect, grounded in how the brain processes and learns from information. When language or other stimuli are completely predictable, there is nothing new to learn. Conversely, when they are entirely random, no meaningful patterns can be discerned. It is in the middle ground—where patterns are discernible but not obvious—that the brain engages most deeply, activating reward mechanisms that make the experience pleasurable. While there is evidence that the U-shaped effect exists across different aesthetic domains, there is also a strong indication that our preferences of aesthetic stimuli are highly individual~\cite{Gold9397}\cite{Althuizen2021}.
Building on these insights, I hypothesize that short lines of text in the mid-region of the random-predictable spectrum are most likely to be useful as seeds for ideas and inspiration during the Seed phase of the creative process. In the next section I introduce two generative AI architectures, LSTM-VAE and LLMs, and in the subsequent section I discuss why LSTM-VAE is more suitable as source of evocative lines during the Seed phase compared to LLMs.

\section{Neural Text Generative Models}
Researchers in Natural Language Processing have worked on text generative models for decades with varying success. The first major breakthrough took place around 2016 with the successful adoption and development of neural network architectures based on Recurrent Neural Networks (RNN), especially their more effective variant, Long Short Term Memory (LSTM) networks~\cite{Hochreiter1997LSTM}. The second major breakthrough came in early 2020s with the introduction of Large Language Models based on the Transformer architecture~\cite{vaswani2017attention}.

It is important to clarify in this context what is a generative natural language model. It is a model that is capable of producing original sequences with or without conditioning in the form of a text (e.g., prompt in an LLM) or any other signal (e.g., image, audio, video). The word ``original'' here is used without any value judgement describing the novelty or aesthetic merit of the text. It simply refers to a text that is not directly derived from the input, e.g., replication of the input or any other hard-coded or rule-based response.

This section compares two neural network architectures for creative text generation: Variational Autoencoders based on Long-Short Term Memory Networks (LSTM-VAE) and Transformer-based Large Language Models (LLMs).

\subsection{LSTM-VAE: Generating Linguistic Variability}
Before the invention of the Transformer-based Large Language Model (LLM), one of the most effective neural architectures for text generation was the Variational Autoencoder (VAE)~\cite{kingma2014auto}. The VAE combined with Long Short-Term Memory networks (LSTM) is commonly referred to as LSTM-VAE. This model is particularly well-suited for generating short, evocative sentences that balance randomness and predictability, making it valuable for creative applications. At its core, the LSTM-VAE consists of two main components:

\textbf{Encoder:} This part of the model reads an input sentence and compresses it into a condensed representation known as the latent vector. Think of this as capturing the essence of the sentence—its underlying semantic, stylistic and syntactic features—in a numerical form. The domain in which latent vectors exist is called the latent space.

\textbf{Decoder:} This component takes the latent vector and attempts to reconstruct the original sentence from it. Essentially, it translates the compressed representation back into natural language.

During training, the model learns to encode sentences into the latent space and then decode them back to their original form. The key innovation of the VAE is that it encourages the latent space to be smooth and continuous. This means that small changes in the latent vector result in small changes in the generated sentence, allowing for controlled variability.

By sampling different points from the latent space, we can generate new, original sentences that are similar but not identical to those in the training data. Because the latent space is structured, sentences generated from nearby points in the space will share certain characteristics, such as themes, syntactic structure or stylistic elements.

After the LSTM-VAE is trained, we can generate new sentences by following these two steps:

\textbf{Sampling:} Randomly selecting a point in the latent space. This introduces an element of randomness, allowing for the creation of novel sentences.

\textbf{Decoding:} Feeding this point into the decoder to generate a sentence. The decoder transforms the latent vector back into a sequence of words, producing a coherent sentence.

Unlike later Large Language Models, LSTM-VAE is only able to generate short sentence-length texts. However, an important advantage of the LSTM-VAE is that it does not require massive amounts of data to produce interesting results. Because it focuses on capturing the essential features of sentences, it can be trained on smaller, curated datasets. This allows for greater control over the style and content of the generated text.

Furthermore, the LSTM-VAE can be extended to include additional information—a variation known as the Conditional Variational Autoencoder (CVAE). For example, we can condition the model on an external signal, such as an audio clip or a specific theme, to guide the generation process~\cite{vechtomova2021lyricjam}. This means the model can produce sentences that are not only original but also relevant to a particular context or artistic intent.

\subsection{Large Language Models (LLMs)}
Large Language Models (LLMs) are currently among the most powerful models for generating text. These models are trained on vast amounts of text data and are designed to predict the next word in a sequence, given the preceding context.

The core component of LLMs is the Transformer architecture, which uses the mechanism of self-attention to process all words in a sentence at once. This allows the model to understand relationships between words, regardless of their position in the sentence, making it highly effective at capturing long-range dependencies in text.

The learning objective of LLMs is typically framed as a language modeling task, where the model learns to predict the probability of the next word in a sequence. For example, given the partial sentence ``The bird sat on the,'' the model would assign probabilities to different possible next words (e.g., ``branch,'' ``perch'') based on patterns it learned during training.

Once trained, LLMs can generate coherent and contextually appropriate text by sampling words one by one, using the previous words as context. Unlike the LSTM-VAE, the Transformer-based LLMs can generate longer texts.

In linguistic terms, LLMs excel at capturing surface-level regularities and common syntactic constructions but can be less effective at generating linguistically novel or unexpected expressions. This predictability can limit their utility in stimulating linguistic creativity, as the generated text may lack the unpredictability to serve as effective creative stimuli.

\subsection{AI-generated Lines as Linguistic Stimuli During the Seed Phase of Creativity}
I contend that LSTM-VAE models are more suitable for generating linguistic stimuli that balance randomness and predictability, making them effective tools during the Seed phase of the creative process. The key reasons are summarized below:

\textbf{Controlled randomness.} LSTM-VAE models introduce randomness through latent space sampling, leading to the generation of linguistically novel sentences that can disrupt habitual language patterns and stimulate creativity. LLMs do have an element of randomness in the sampling of the next token, which is controlled by the temperature parameter. Higher temperature values make the model more likely to choose less probable words, while lower values make it more conservative. However, this randomness is limited as the model only samples from among the tokens with the highest probabilities in each position.

\textbf{Syntactic Diversity and Defamiliarization.} The latent representations in LSTM-VAE models capture a wide range of syntactic variations, allowing for the generation of syntactically diverse and syntactically defamiliarized lines. LSTM-VAE samples the latent vector representing the entire sentence from the latent space, which opens the possibility of generating dramatically different sentences syntactically and stylistically.

\textbf{Limited Memorization.} Due to their smaller size and training objectives, LSTM-VAE models are less prone to memorizing training data, reducing the likelihood of generating clichéd or overly predictable language. Moreover, the training dataset can be curated, for example, only allowing sources that conform to the desired artistic style or genre. This kind of careful curation is not feasible with LLMs, as they require training data that is orders of magnitude larger to effectively train their billions of parameters.

Transformer-based LLMs prioritize coherence and adherence to learned linguistic patterns, which may result in text that is less effective in disrupting linguistic expectations and stimulating creativity.

\section{Comparative Analysis of AI-Generated Lines as Sources of Artistic Inspiration}
In this section, I analyze the characteristics of lines generated by two different AI systems—LSTM-VAE and LLM—focusing on their potential to serve as seeds for artistic inspiration. The analysis compares two poems (Figure~\ref{fig:poems}). The first poem~\cite{vechtomova2018echoes} (Figure 1a) emerged through the author's creative engagement with a pool of 300-400 lines generated by an LSTM-VAE model~\cite{vechtomova2018generatinglyrics}. Rather than following a predetermined structure or narrative, the creative process involved first selecting lines that resonated emotionally, then discovering connections and meaning through their arrangement. In contrast, the second poem (Figure 1b) was generated autonomously by ChatGPT-4o in response to a prompt requesting original lyrics.

\begin{figure}[htbp]
\centering
\begin{minipage}[t]{0.50\textwidth}
\centering
\textbf{(a)}
\begin{verse}
With a shadow beside, \\
All the tears inside \\
I turned back to the earth. \\
She walks in a better moon \\
And I let my heart go. \\
You'll be with freedom and shelter \\
Begin to the next mother earth. \\
And the light shines on the runway \\
Driving an endless ship on fire. \\
When I'm frightened in the air \\
Rooted in the light \\
There's the garden in the darkness for us. \\
When the promise in the rain \\
And the stars they go.
\end{verse}
\end{minipage}
\hfill
\begin{minipage}[t]{0.48\textwidth}
\centering
\textbf{(b)}
\begin{verse}
In the twilight of a memory \\
Where the shadows gently sway \\
I saw knights of old, with tales untold, \\
And dreams that led the way \\
There were voices calling in the night \\
Whispers lost in the breeze \\
Echoes of our fragile Earth \\
And the secrets in the trees \\
I was standing in the ruins of a promise \\
With a midnight sky above \\
Searching for a sign, a spark divine \\
In a world that's lost its love \\
There were visions dancing in my mind \\
Of a time when hearts were free \\
I was yearning for a moment's peace \\
In a song of mystery
\end{verse}
\end{minipage}
\caption{Poems: (a) Poem \#1 composed by the author from lines generated by LSTM-VAE; (b) Poem \#2 generated autonomously by LLM (ChatGPT-4o).}
\label{fig:poems}
\end{figure}

\subsection{LSTM-VAE Generated Lines}
The LSTM-VAE system produces lines with several qualities that make them particularly effective as creative stimuli:

\textbf{Semantic Openness.} The lines often possess an intentional incompleteness or ambiguity that invites completion by the artist's imagination: "with a shadow beside", "all the tears inside", "rooted in the light". Such semantic openness works not through randomness but through carefully balanced indeterminacy. Each line provides enough concrete imagery or emotional resonance to anchor meaning, while remaining open enough to invite multiple interpretations.

\textbf{Emotional Resonance without Explicit Statement.} Rather than directly stating emotions, the lines create emotional spaces through suggestive imagery: ``when I'm frightened in the air'', ``and I let my heart go'', ``she walks in a better moon''. These lines evoke emotional states while leaving their specific context undefined, allowing artists to project their own experiences and interpretations.

\textbf{Unconventional Juxtapositions.} The system creates unexpected combinations that can spark new creative directions: ``driving an endless ship on fire'', ``rooted in the light''. These combinations challenge conventional associations while remaining emotionally evocative. The first line creates an unusual and dramatic imagery, which defies immediate resolution as it requires the reader to reconcile paradoxical combinations, such as "endless ship". The second line ``rooted in the light'' is evocative because it combines two concepts with different associations: ``rooted'' is associated with stability and permanence, while ``light'' is associated with something fleeting and ephemeral.

\textbf{Syntactic Estrangement and Incompleteness.} The lines exhibit both deliberate syntactic incompleteness and semantic-syntactic defamiliarization. Many lines have an open-ended quality that invites completion or continuation: ``when the promise in the rain'', ``and the stars they go''. Such syntactic fragments do more than invite completion—they exemplify how linguistic estrangement can generate new poetic possibilities. When the text presents constructions like "walks in a better moon" or ``begin to the next mother earth,'' it isn't simply varying its structural patterns. Rather, these disruptions of expected syntactic-semantic relationships create what we might call ``productive ungrammaticality.'' The seeming ``errors'' (like the doubled subject in ``the stars they go'') function as deliberate poetic devices that slow down perception and force fresh engagement with language. These moments of syntactic estrangement transform conventional linguistic patterns into spaces of semantic possibility, where meaning emerges not from standard grammatical relationships but from more fluid, associative connections. In this way, the text's syntactic disruptions serve the same purpose as Shklovsky's ostranenie—making the familiar strange to heighten perception and unlock new modes of understanding.

\subsection{LLM-Generated Lines}
The LLM produces lines that demonstrate different characteristics:

\textbf{Syntactic and Semantic Completion.} Each line demonstrates technical competence through complete but conventional syntactic-semantic relationships: ``where the shadows gently sway'' (adverbial clause with standard modifier-verb structure), ``i saw knights of old, with tales untold'' (transitive construction with parallel prepositional phrases).

While technically accomplished in their varied structures, these lines operate within expected linguistic patterns. Their adherence to conventional syntax, while creating polished poetry, actually constrains their potential for defamiliarization. Unlike texts that employ productive syntactic disruption, these complete syntactic units leave less room for the kind of creative reinterpretation that Shklovsky identified as essential to artistic perception.

\textbf{Conventional Imagery and Tropes.} The lines rely heavily on established poetic imagery: ``whispers lost in the breeze'', ``with a midnight sky above'', ``and dreams that led the way''. While evocative, these images draw from familiar poetic vocabulary rather than creating new conceptual spaces. 

\textbf{Narrative Coherence.} The lines are constructed to fit into a clear narrative framework: ``i was standing in the ruins of a promise'', ``searching for a sign, a spark divine''. This narrative clarity, while aesthetically pleasing, may limit their ability to open unexpected creative directions.

\textbf{Traditional Poetic Devices.} The lines consistently employ conventional poetic techniques, such as hyperbaton (``spark divine'', ``tales untold''), metaphor (``twilight of a memory'') and regular meter and rhyme. While skillfully crafted, this adherence to traditional poetic forms may constrain rather than stimulate creative exploration.

\subsection{From Inspiration to Creation}
Poem \#1 (Figure 1a) demonstrates how LSTM-VAE generated lines can serve as building blocks for original creative work. Rather than selecting lines to fit a preconceived narrative, the author's emotional resonance with certain lines guided the initial selection, after which a deeply personal narrative emerged organically through the process of engaging with and arranging these fragments. What began as an intuitive response to evocative imagery and emotional undertones gradually crystallized into a story of departure, loss, and memory. The progression from vulnerability to transcendent resolution was not planned, but rather discovered through the act of composition itself. While this emergent narrative holds profound personal meaning for the author, the poem's suggestive imagery and emotional depth invite readers to form their own interpretations, demonstrating how AI-generated fragments can inspire authentic artistic creation without constraining it to predetermined paths. The poem's interpretive nature is possible due to several qualities of the LSTM-VAE lines:

\begin{itemize}

\item Their syntactic estrangement (``when the promise in the rain'') creates productive disruptions that enable novel semantic connections.
\item The paradoxical imagery (``rooted in the light'') creates a strong emotional effect as they are not immediately recognizable.
\item The semantic-syntactic tensions in certain images (``she walks in a better moon'', ``driving an endless ship on fire'') open spaces for metaphorical development.
\item The deliberate incompleteness of some lines (``with a shadow beside'') invites readers to resolve their inherent indeterminacy.
\end{itemize}

This case study suggests that the very qualities that make LSTM-VAE lines seem less ``polished''—their ambiguity, incompleteness, and unconventional combinations—may make them more valuable as seeds for artistic inspiration. While LLM produces more technically accomplished lines, their completeness and adherence to convention may make them less effective as creative stimuli. This suggests that in developing AI systems as tools for artistic inspiration, we should focus less on traditional metrics of poetic quality and more on qualities that create space for human creative interpretation and transformation.

\section{Conclusion}
This paper set out to explore the characteristics of AI-generated poetic lines as possible stimuli for artistic inspiration during the Seed phase of the creative process. For decades, artists, writers and poets have used methods like the cut-up technique and deliberate misuse of language to break from conventional norms and habitual uses of language. While these techniques introduced elements of randomness, they often worked within aesthetic constraints. As Nees~\cite{nees1969generative} observed, randomness alone cannot create art without an aesthetic structure to guide it. This insight is supported by studies in the psychology of aesthetics, which suggest the existence of the inverted U-shaped (Wundt) effect, according to which artifacts of intermediate complexity are the most stimulating and aesthetically pleasing.

As neural generative models are increasingly used by people for a variety of creative tasks, they have the potential to generate stimuli that might be conducive to inspiring the artistic process. I performed a comparative analysis of two different approaches to AI-generated poetry: (a) individual lines generated by an LSTM-VAE system that were later curated by a human author into a complete poem, and (b) a complete poem autonomously generated by an LLM. This analysis revealed fundamental differences in how these systems generate content that might serve as creative inspiration.

The LSTM-VAE generated lines demonstrate syntactic defamiliarization and indeterminacy, with their very incompleteness and ambiguity serving as stimuli for creative discovery. This is evidenced in how meanings and narratives emerged organically through the process of engaging with these lines, resulting in an original poem that transcends its individual components. The resulting work demonstrates how the semantic openness and emotional resonance of the LSTM-VAE lines enabled not just creative transformation, but the forming of an emotionally authentic artistic narrative.

In contrast, while an LLM produces technically accomplished and aesthetically pleasing poetry, its output tends toward conventional imagery and complete poetic units that leave less room for creative interpretation. The system's adherence to traditional poetic devices and narrative coherence is impressive from a technical standpoint, but may limit its usefulness as a source of creative inspiration.

This distinction aligns with Pepperell's~\cite{pepperell2007art} observations about indeterminacy, which causes our habitual recognition to be suspended as the mind tries to assign meaning to the indeterminate stimulus. This is consistent with Gold's~\cite{Gold9397} explanation of the Wundt effect, who states that the mind engages more meaningfully with stimuli in the intermediate range between randomness and predictability. The LSTM-VAE generated lines, with their combination of structural complexity and semantic openness, appear to occupy this fertile middle ground between chaos and convention—a space where artistic meaning can emerge through engaged creative exploration.

I contend that it is not sufficient to produce technically accomplished verse or to create superficial combinations of unrelated concepts. To serve as effective stimuli for artistic inspiration, generated lines should possess qualities that create space for creative discovery—emotional resonance without explicit statement, syntactic openness that invites interpretation, and unconventional juxtapositions that spark new associations. The organic emergence of narrative through engagement with LSTM-VAE generated lines demonstrates how these characteristics can enable artistic creation.

This suggests that in developing AI systems as tools for artistic inspiration, we should focus less on traditional metrics of poetic quality and more on generating output that occupies the fertile middle ground between randomness and predictability. The goal should be to create systems that generate truly evocative seeds—unexpected yet meaningful, unresolved yet emotionally resonant—inviting artists to discover the narratives and meanings that emerge through deep creative engagement with these fragments.

\bibliographystyle{unsrt}  
\bibliography{lacus}

\end{document}